\documentclass{article}

\usepackage{arxiv}

\usepackage[utf8]{inputenc} 
\usepackage[T1]{fontenc}    
\usepackage{hyperref}       
\usepackage{url}            
\usepackage{booktabs}       
\usepackage{amsfonts}       
\usepackage{nicefrac}       
\usepackage{microtype}      
\usepackage{lipsum}

\usepackage{bm}
\usepackage{amsmath}
\usepackage{amssymb}
\usepackage{amsfonts}
\usepackage{graphicx}
\def\vec#1{\mathbf{#1}}
\usepackage{algorithm}
\usepackage{algorithmic}
\usepackage{comment}

\DeclareMathOperator*{\argmin}{arg\,min}
\newcommand{\indep}{\mathop{\perp\!\!\!\!\perp}}

\title{Meta-learning for heterogeneous treatment effect estimation with closed-form solvers}

\author{
  Tomoharu Iwata\\
  NTT Communication Science Laboratories\\
  \And
  Yoichi Chikahara\\
  NTT Communication Science Laboratories\\
}
\date{}

\begin{document}
\maketitle

\begin{abstract}
This article proposes a meta-learning method for estimating the conditional average treatment effect (CATE) from a few observational data. The proposed method learns how to estimate CATEs from multiple tasks and uses the knowledge for unseen tasks. In the proposed method, based on the meta-learner framework, we decompose the CATE estimation problem into sub-problems. For each sub-problem, we formulate our estimation models using neural networks with task-shared and task-specific parameters. With our formulation, we can obtain optimal task-specific parameters in a closed form that are differentiable with respect to task-shared parameters, making it possible to perform effective meta-learning. The task-shared parameters are trained such that the expected CATE estimation performance in few-shot settings is improved by minimizing the difference between a CATE estimated with a large amount of data and one estimated with just a few data. Our experimental results demonstrate that our method outperforms the existing meta-learning approaches and CATE estimation methods.
\end{abstract}

\section{Introduction}
\label{sec:introduction}

Treatment effects are often heterogeneous across individuals.
For instance, the effects of medical treatment (e.g., drug administration) on health status differ across patients,
and those of online advertisement on purchase decision
vary depending on consumers.
In recent years, there has been a growing interest 
in the estimation of such heterogeneous treatment effects
as it allows us to make data-driven decision making in various fields,  
such as precision medicine~\cite{bica2019estimating,gao2021assessment,shalit2020can},
personalized advertisement~\cite{bottou2013counterfactual,wang2015robust,sun2015causal,zou2020counterfactual},
and educational program design~\cite{xie2020heterogeneous}.

To measure the heterogeneity in treatment effects,
existing methods use a conditional average treatment effect (CATE),
which is defined as an average treatment effect across individuals
with identical feature attributes.
To accurately estimate the CATE, many of these methods fit complex machine learning models to large-scale observational data~\cite{shalit2017estimating,yoon2018ganite,wager2018estimation,kunzel2019metalearners,curth2021nonparametric}.
However, preparing such large-scale observational data can be difficult.
In the case of medical treatment, for example,
there are many small hospitals (with fewer than 100 beds~\cite{wiens2014study}), 
and each hospital has a different patient population
in feature attributes, outcome, and treatment preference.
To widen the scope of applications,
it is crucial to develop a strategy for estimating a CATE from
a small observational dataset by effectively leveraging the information from other datasets
while taking into account the population difference.

This article proposes such a CATE estimation strategy based on meta-learning,
which has been successfully used in classification and regression tasks~\cite{finn2017model}.
The goal of meta-learning is to improve the test performance 
when there are only a few data in a test task.
The key idea is to use the data from various tasks to extract 
the \textit{task-shared} knowledge,
which serves as the prior knowledge that is helpful for unseen tasks.
This idea is also practical for the CATE estimation tasks.
To see this, revisit the aforementioned medical treatment example 
and consider the tasks of estimating the CATE from patient records for each hospital.
To build a CATE estimation model for unseen small hospitals with only a few patient records,
we need to effectively extract prior knowledge from the records from many other hospitals,
which is exactly the above idea of meta-learning.

\begin{figure*}[t!]
  \centering
  \includegraphics[width=35em]{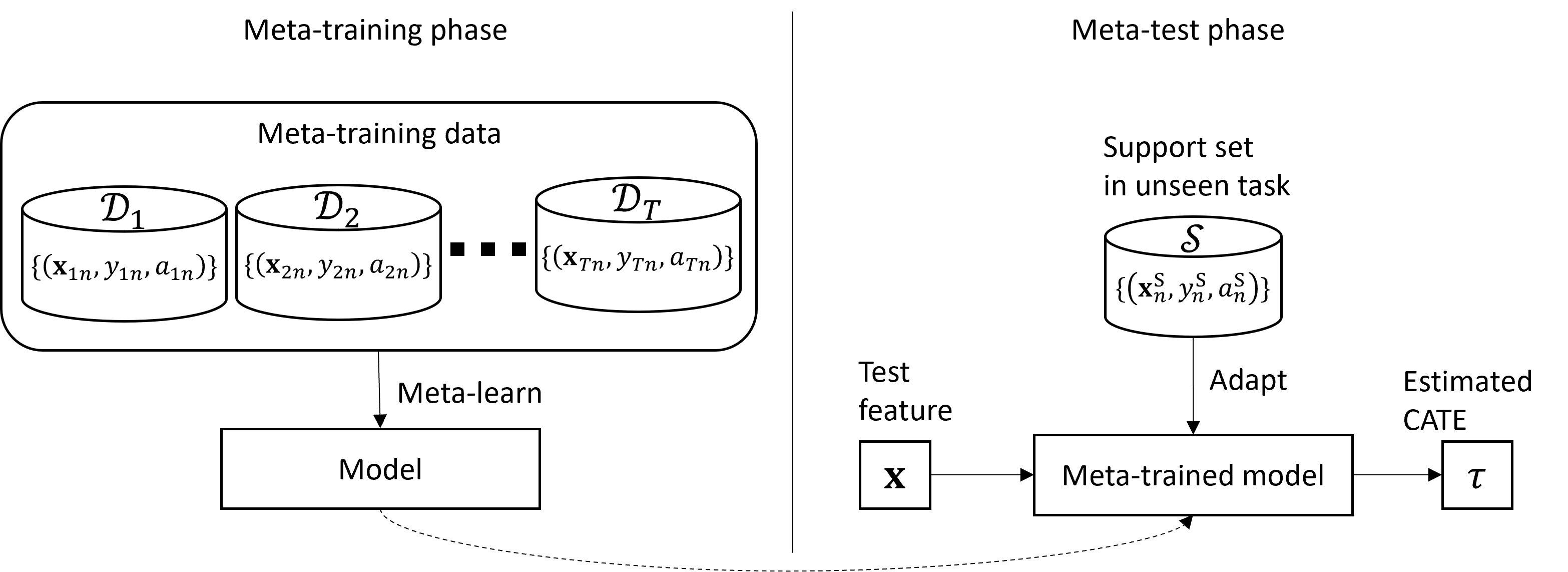}
  \caption{Our problem setup. In the meta-training phase, we are given meta-training data from multiple tasks for meta-learning our model. In the meta-test phase, we are given a few data, which is called a support set, in an unseen task. Using the meta-trained model adapted by the support set, we estimate CATE of a test individual given its feature attributes.}
  \label{fig:setting}
\end{figure*}

Following this idea, as illustrated in Figure~\ref{fig:setting},
we consider the following meta-learning problem setup.
Suppose that we are given multiple meta-training tasks, 
each of which contains a large number of observational data on
the features, outcomes, and treatments of individuals.
Here a task refers to a CATE estimation task on a treatment.
Our goal is to improve the CATE estimation performance on unseen tasks with only a few data,
where
the population in the unseen tasks is different from the meta-training tasks,
but related to some of the meta-training tasks. 

Developing a meta-learning approach for CATE estimation faces
two main challenges.
The first challenge is computational complexity for solving the bi-level optimization problem for meta-learning~\cite{huisman2021survey}.
This bi-level problem consists of the inner and outer optimization problems:
the former adapts task-specific parameters to each task,
and the latter finds task-shared parameters
that maximize the expected test performance when adapted to each task.
The computational difficulty arises
because we need to repeatedly solve the inner optimization problem
for each iteration of the outer optimization.
A standard approach for solving this inner problem is to perform iterative optimization
with a gradient descent method~\cite{finn2017model}.
However, since this approach requires a large number of iterations whenever adapting to each task,
it is computationally expensive to solve the overall bi-level optimization problem.
To overcome this computational challenge, 
we formulate a CATE estimation model so that we can analytically solve the inner optimization problem
in a closed form that is differentiable with respect to task-shared parameters.
In particular, our model is based on the CATE estimation framework 
called a \textit{meta-learner}~\cite{kunzel2019metalearners},
which decomposes the CATE estimation problem into sub-problems
and solves each sub-problem using machine learning models called \textit{base learners}.
As base learners, we use neural networks with task-shared and task-specific parameters 
in which the optimal task-specific parameters are obtained in a differentiable closed form for each sub-problem.
Then, we can backpropagate the loss through the sub-problems to update the task-shared parameters for all sub-problems.
In Sections~\ref{sec:task-adapted} and \ref{sec:solver},
we present a formulation example with a doubly-robust learner (DR-Learner)~\cite{kennedy2020optimal},
which is a successfully used meta-learner that solves a two-stage regression problem~\cite{curth2021nonparametric}.

The second challenge is that naively applying the existing meta-learning methods
for classification and regression 
does not necessarily lead to good CATE estimation performance.
Achieving good estimation performance with such naive methods requires 
to measure the loss using the true CATE values.
Unfortunately, however, we have no access to such true CATE values.
This is because a treatment effect is defined as a difference between two \textit{potential outcomes}
(i.e., the outcomes when an individual is treated and when not treated),
which can never be jointly observed for each individual.
To resolve this issue,
as alternatives to the true CATE values,
in the meta-training phase,
we construct the \textit{pseudo CATEs} by employing the estimation model fitted 
to large data in each meta-training task.
By minimizing the difference between these pseudo CATEs 
and the one estimated from small data in meta-training tasks,
we perform meta-learning such that the expected CATE estimation performance is directly improved.

Our contributions are summarized as follows:
\begin{enumerate}
  \item We propose a meta-learning method that improves the
    test performance of estimating CATE given a limited amount of observational data 
    by minimizing the difference between a CATE estimated with large data and
    one estimated with small data in various tasks.
  \item We develop a differentiable closed-form solver to adapt the CATE model to each task for effective meta-learning, where we decompose the CATE estimation problem into sub-problems that can be solved analytically.
  \item We empirically demonstrate that our proposed method achieves good CATE estimation performance in few-shot settings, where we have access to only a few instances of observational data, compared with the existing meta-learning methods.
\end{enumerate}

\section{Related work}

A large number of studies have been dedicated to overcoming the sample selection bias induced 
by the so-called fundamental problem of causal inference~\cite{holland1986statistics},
namely, that we can never jointly observe the two potential outcomes for each individual
(see e.g., \cite{yao2021survey} for a literature review).
A major classical strategy for dealing with this sample selection bias
is matching~\cite{rubin1973matching,abadie2006large},
which aims to estimate the unobserved potential outcome 
by employing the information about the neighbor individual
who is nearest to the target individual in terms of feature attributes.
Although seeking the nearest neighbor is difficult especially in a high-dimensional setting,
the traditional methods tackle this difficulty 
by seeking the neighbors in the space of \textit{propensity score},
which is the probability that an individual receives a treatment~\cite{rosenbaum1983central,rosenbaum1985constructing}.

The recent methods have achieved successful CATE estimation performance
using complex machine learning models, such as
tree-based models~\cite{wager2018estimation,wang2015robust},
Gaussian processes~\cite{alaa2017bayesian},
and neural networks~\cite{johansson2016learning,shalit2017estimating,yoon2018ganite}.
Among these methods,
Counterfactual regression (CFR)~\cite{shalit2017estimating}
is a well-known neural-network-based method,
which learns a shared representation that is used
to predict two potential outcomes.
In contrast to these model-specific methods,
the model-agnostic methods,
such as T-Learner, X-Learner, and DR-Learner~\cite{kunzel2019metalearners,kennedy2020optimal},
can use any arbitrary machine learning model
and hence have attracted a growing attention
as a flexible tool for investigating heterogeneous treatment effects.
Note that although these methods are called meta-learners,
they are \textbf{not} related to meta-learning (i.e., learning to learn)
in the field of machine learning.
In this article, we use ``meta-learning'' to denote learning to learn,
which uses the data from various tasks~\cite{finn2017model}.

The weakness of these machine-learning-based methods is that 
they require a lot of observational data for model fitting.
To resolve this weakness,
many methods aim to improve the data efficiency
by employing the techniques from 
transfer learning~\cite{kunzel2018transfer,johansson2018learning,shi2021invariant,bicatransfer2022}
and multi-task learning~\cite{alaa2017bayesian,zhu2021direct,kyono2021selecting,chu2022multi}.
A common disadvantage of these methods is that 
they require data about target tasks in the training phase,
which limits the scope of the CATE estimation applications.
Moreover,
transfer-learning-based methods assume that there are only two tasks (i.e., source and target),
and multi-task-learning-based methods train the models without consideration of the few-shot setting.
Due to these disadvantages, none of these methods 
can be applied to the real-world scenarios
where we have access to only a few instances of observational data.
To overcome these weaknesses,
the proposed meta-learning-based method trains the models 
such that the test performance given a small number of observational data is improved.
To perform such a training procedure,
it aims to simulate the few-shot setting in meta-training steps
based on an episodic training framework~\cite{vinyals2016matching}.

Although there are several causal inference methods founded on meta-learning,
their problem setup is different;
most of these methods focus on causal discovery
(i.e., the inference of the causal directions among random variables)~\cite{bengio2019meta,ke2019learning,ton2021meta},
not on the CATE estimation.
A notable exception is the Meta-CI method~\cite{sharma2019metaci},
which borrows the idea of the CFR method~\cite{shalit2017estimating} and 
learns a shared representation for each task to estimate the two potential outcomes.
Compared with Meta-CI, our proposed method has two advantages.
First, it performs meta-learning such that the expected CATE estimation
performance is improved
using the CATE values estimated with large data as supervision.
Utilizing such CATE values, which we call pseudo CATEs, is important 
for achieving high performance in CATE estimation,
where we have no access to the true labels.
The second advantage is that it offers closed-form task adaptation for CATE estimation.
This advantage is crucial in meta-learning
because
the recent studies
show that the closed-form solvers for task adaptation
can greatly improve the classification and
regression performances~\cite{snell2017prototypical,bertinetto2018meta}.
Whereas there are several closed-form solvers 
in the field of meta-learning~\cite{snell2017prototypical,bertinetto2018meta,patacchiola2020bayesian,iwata2022few},
none of them can be directly applied to training the models in meta-learners,
which are founded on the two-stage-based estimation,
where the outputs of one regression model are used for training another regression model.
Although developing closed-form solvers in such a setting is challenging,
it makes a significant contrast to the meta-CI method,
which, as with CFR, is founded on a nonconvex regularizer on distributional discrepancy,
making impossible to analytically solve the optimization problem.
In Section~\ref{sec:experiments},
we experimentally show that each of these advantages yields a considerable estimation performance gain.

\section{Proposed method}
\label{sec:proposed}

\begin{figure*}[t!]
  \centering
  \includegraphics[width=35em]{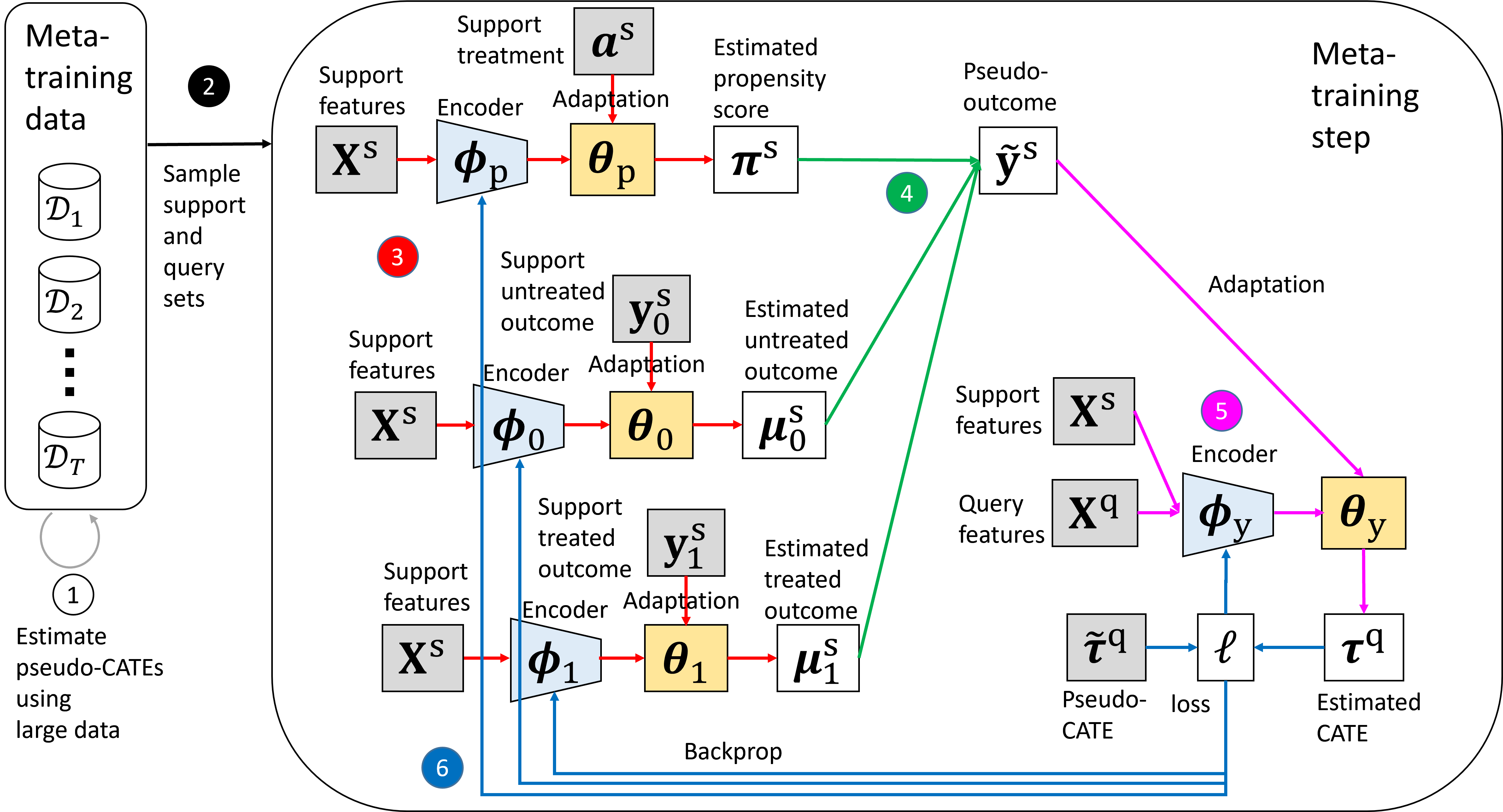}
  \caption{Our meta-learning framework:
    \textbf{1)}~At the beginning of meta-learning phase, pseudo CATEs $\{\tilde{\tau}_{n}\}_{n=1}^{N_{t}}$,
    which are used as pseudo labels for the true CATE values,
    are estimated using large data in each task $\mathcal{D}_{t}$.
    \textbf{2)}~For each meta-learning step, support set $\{(\vec{x}_{n}^{\rm{s}},y_{n}^{\rm{s}},a_{n}^{\rm{s}})\}_{n=1}^{N_{\rm{s}}}$ and query set $\{(\vec{x}_{n}^{\rm{s}},\tilde{\tau}_{n}^{\rm{s}})\}_{n=1}^{N_{\rm{q}}}$ are randomly sampled from a task in meta-training data (black arrow).
    \textbf{3)}~For all support instances, propensity scores $\bm{\pi}^{\rm{s}}=\{\pi(\vec{x}^{\rm{s}}_{n};\hat{\bm{\theta}}_{\rm{p}},\bm{\phi}_{\rm{p}})\}_{n=1}^{N_{\rm{s}}}$, outcomes when untreated $\bm{\mu}^{\rm{s}}_{0}=\{\mu(\vec{x}^{\rm{s}}_{n};\hat{\bm{\theta}}_{0},\bm{\phi}_{0})\}_{n=1}^{N_{\rm{s}}}$, and outcomes when treated $\bm{\mu}^{\rm{s}}_{1}=\{\mu(\vec{x}^{\rm{s}}_{n};\hat{\bm{\theta}}_{1},\bm{\phi}_{1})\}_{n=1}^{N_{\rm{s}}}$ are estimated using task-shared encoders with parameters $\bm{\phi}_{\rm{p}}, \bm{\phi}_{0}, \bm{\phi}_{1}$ and task-specific models with parameters $\hat{\bm{\theta}}_{\rm{p}},\hat{\bm{\theta}}_{0},\hat{\bm{\theta}}_{1}$ (red arrows). Here task-specific parameters are adapted by support treatments $\vec{a}^{\rm{s}}=\{a_{n}^{\rm{s}}\}_{n=1}^{N_{\rm{s}}}$, support untreated outcomes $\vec{y}_{0}^{\rm{s}}=\{y_{n}^{\rm{s}}\}_{n:a_{n}^{\rm{s}}=0}$, and support treated outcomes $\vec{y}_{1}^{\rm{s}}=\{y_{n}^{\rm{s}}\}_{n:a_{n}^{\rm{s}}=1}$.
    \textbf{4)}~For all support instances, pseudo outcomes $\tilde{\vec{y}}^{\rm{s}}=\{\tilde{y}_{n}^{\rm{s}}\}_{n=1}^{N_{\rm{s}}}$ are computed by Eq.~(\ref{eq:tilde_y}) (green arrows).
    \textbf{5)}~For all query instances, CATEs $\bm{\tau}^{\rm{q}}=\{\tau(\vec{x}_{n}^{\rm{q}};\hat{\bm{\theta}}_{\rm{y}},\bm{\phi}_{\rm{y}})\}_{n=1}^{N_{\rm{q}}}$ are estimated using a task-shared encoder with parameters
    $\bm{\phi}_{\rm{y}}$ and a task-specific model with parameters $\hat{\bm{\theta}}_{\rm{y}}$; latter is adapted with pseudo outcomes $\tilde{\vec{y}}^{\rm{s}}$ (purple arrows).
    \textbf{6)}~Loss $\ell$ between pseudo CATEs $\tilde{\bm{\tau}}^{\rm{q}}=\{\tilde{\tau}_{n}^{\rm{q}}\}_{n=1}^{N_{\rm{q}}}$
    and estimated CATEs $\bm{\tau}^{\rm{q}}$ is backpropagated to update encoders $\bm{\phi}_{\rm{p}}, \bm{\phi}_{0}, \bm{\phi}_{1}, \bm{\phi}_{\rm{y}}$ (blue arrows).
    Observed and estimated variables are shown in gray and white rectangles, respectively. Task-specific parameters and task-shared encoders are displayed in orange rectangles and light-blue trapezoids, respectively.}
  \label{fig:framework}
\end{figure*}

This section is organized as follows.
In Section~\ref{sec:problem}, we present our problem setup.
In Section~\ref{sec:task-adapted}, we provide a general formulation of the task-specific CATE models,
which are adapted to the given small observational data for each task.
In Section~\ref{sec:solver}, we propose a specific model formulation 
that enables us to obtain the optimal model parameters for task adaptation
in a differentiable closed form.
In Section~\ref{sec:meta-learn}, we present
our meta-learning algorithm
for training the task-shared parameters in our models such that
the expected CATE estimation performance is improved.
Figure~\ref{fig:framework} illustrates an overview of
our proposed meta-learning framework.
In Section~\ref{sec:variant}, we present variants of the proposed method.

\subsection{Problem setup}
\label{sec:problem}

We consider the CATE estimation task, i.e., the task of 
 predicting the average effect of treatment $A$
on outcome $Y$ in a subgroup of individuals who have identical
feature attributes $\vec{X} = \vec{x}$.
Here we consider binary treatment $A\in\{0,1\}$,
which takes $A=1$ if an individual is treated and $A=0$ otherwise,
continuous-valued outcome $Y\in\mathbb{R}$,
whereas each element in feature vector $\vec{X}\in\mathcal{X}$
takes a discrete or continuous value.
Under the binary treatment setup,
CATE is defined as
$\mathbb{E}[Y(1)-Y(0)|\vec{X}=\vec{x}]$,
where $\mathbb{E}$ denotes the expectation,
and $Y(0)$ and $Y(1)$ respectively denote random variables called
potential outcomes, each of which represents the outcome when $A=0$ and $A=1$~\cite{rubin1974estimating}.

In this article, we consider the meta-learning problem for the CATE estimation task:
given many CATE estimation tasks in the meta-training phase,
we aim to improve the CATE estimation performance on unseen meta-test tasks,
where there are only a few instances of observational data.
Formally, this meta-learning problem can be described as follows. 
In the meta-training phase, we are given $T$ CATE estimation tasks.
The input observational data of each task $t\in\{1,\cdots,T\}$ are given as
$\mathcal{D}_{t}=\{(\vec{x}_{tn},y_{tn},a_{tn})\}_{n=1}^{N_{t}} \sim \mathrm{P}^{(t)}(\vec{X}, Y, A)$
i.e., a set of i.i.d. observations of 
feature vector $\vec{x}_{tn}\in\mathcal{X}$, 
outcome $y_{tn}\in\mathbb{R}$, 
and treatment $a_{tn}\in\{0,1\}$,
which are drawn from joint distribution $\mathrm{P}^{(t)}(\vec{X}, Y, A)$.
Using such meta-training data $\mathcal{D}_{1},\dots,\mathcal{D}_{T}$,
we aim to improve the performance on unseen tasks in the meta-test phase.
Unlike the setups of transfer learning and multi-task learning,
the data for these test tasks are not given in the training phase,
and their sample size is exceedingly small.


As with the standard meta-learning methods~\cite{finn2017model},
we consider two data subsets in each task, called a \textit{support set} and a \textit{query set},
each of which is respectively used
to adapt the estimation models to each task
and to evaluate their test estimation performance.
Support set $\mathcal{S}$ is a small data subset in each task,
which is denoted by
$\mathcal{S}=\{(\vec{x}^{\rm{s}}_{n},y^{\rm{s}}_{n},a^{\rm{s}}_{n})\}_{n=1}^{N^{\rm{s}}}$.
We prepare a support set for each meta-training and meta-test task:
in the meta-training phase, the support set is randomly sampled from 
the meta-training data (see Section~\ref{sec:meta-learn} for details),
while in the meta-test phase, it is the whole observational data in each meta-test task.
Support set size $N^{\rm{s}}$ in the meta-training task is a hyperparameter,
which needs to be set to be similar to that in meta-test data.
The support set size is exceedingly smaller than the total data sample size
of each meta-training task, $N_{t}$.
For instance in our experiments,
we used the meta-training data with sample size $N_{t}\in\{747,10000\}$
and set the support set size to $N^{\mathrm{s}}\in\{6,8,14\}$.
Query set $\mathcal{Q}$ is randomly sampled from the meta-training data
in the meta-training phase, which is used to compute the test loss.
In our experiments, its size, $N^{\mathrm{q}}$, is set to 40.

As with many existing CATE estimation
methods~\cite{alaa2017bayesian,johansson2016learning,shalit2017estimating,sharma2019metaci},
we make two standard assumptions.
One is {\it strong ignorability},
which imposes the conditional independence relation between potential outcomes and treatment 
conditioned on a feature vector,
i.e., $Y(0),Y(1)\indep A \mid \vec{X}$.
Roughly speaking, this relation requires feature vector $\vec{X}$ to include all the confounders,
i.e., the variables that influence both treatment $A$ and outcome $Y$,
and hence assumes that there are no unobservable confounders.
The other is {\it positivity}, which ensures
that conditioned on any feature vector value $\vec{x}\in\mathcal{X}$,
the probability of receiving treatment assignment $A=a$ is nonzero
for all $a\in\{0,1\}$; formally
$0 < \mathrm{P}(A=1\mid\vec{X}=\vec{x}) < 1$ for any $\vec{x}\in\mathcal{X}$.
We make these two assumptions for each task $t$
to express the CATE of each task $t$ as a difference between conditional expected values:
\begin{align*}
  \mathbb{E}^{(t)}[Y(1)-Y(0)|\vec{X}=\vec{x}] = \mathbb{E}^{(t)}[Y| \vec{X}=\vec{x}, A=1] - \mathbb{E}^{(t)}[Y| \vec{X}=\vec{x}, A=0],
\end{align*}
where $\mathbb{E}^{(t)}$ denotes the expectation with respect to the distribution of task $t$.
Thus, each CATE estimation task $t$ can be reduced to
estimating the difference, $\mathbb{E}^{(t)}[Y| \vec{X}=\vec{x}, A=1] - \mathbb{E}^{(t)}[Y| \vec{X}=\vec{x}, A=0]$,
from the input i.i.d. observations, $\mathcal{D}_t \sim \mathrm{P}^{(t)}(\vec{X}, Y, A)$.
Additionally, we assume that such a task is drawn from the same distribution between the meta-training and meta-test phases; this assumption is widely used in meta-learning~\cite{finn2017model} to guarantee the generalization performance.

\subsection{Adaptation to CATE estimation task}
\label{sec:task-adapted}

In this section, we explain how we perform task adaptation
using support set 
$\mathcal{S}=\{(\vec{x}^{\rm{s}}_{n},y^{\rm{s}}_{n},a^{\rm{s}}_{n})\}_{n=1}^{N^{\rm{s}}}$
in each meta-training and meta-test task.

The goal of this task adaptation is to estimate the CATE of each task $t$, 
i.e., $\mathbb{E}^{(t)}[Y(1)-Y(0)|\vec{X}=\vec{x}]$, 
where $\mathbb{E}^{(t)}$ denotes the expectation with respect to the distribution of task $t$.
To achieve this goal, we use two sets of model parameters, 
task-shared parameters $\bm{\phi}_{\rm{y}}$ and task-specific parameters $\bm{\theta}_{\rm{y}}^{(t)}$,
each of which is respectively used 
to store the knowledge that is common across tasks
and to handle the task heterogeneity.
With these parameters, we estimate the CATE of each task $t$
as follows:
\begin{align}
  \tau(\vec{x};\bm{\theta}_{\rm{y}}^{(t)},\bm{\phi}_{\rm{y}})\approx\mathbb{E}^{(t)}[Y(1)-Y(0)|\vec{X}=\vec{x}].
  \label{eq:tau}
\end{align}

In meta-learning, such task-shared and task-specific parameters are learned by solving a bi-level optimization problem, whose outer and inner optimization problems are designed 
to learn the task-shared parameters and the task-specific parameters, respectively.
As already described in Section~\ref{sec:introduction},
solving the inner optimization problem for task adaptation with a gradient descent method
needs a large number of iterations and hence requires much computation time.
To efficiently solve this bi-level optimization problem,
it is crucial to obtain the optimal values of task-specific parameters in a closed form.



In this article, we show that we can achieve this goal
by following the meta-learner framework~\cite{kunzel2019metalearners}
and adopting an effective formulation choice for this framework.
In particular, although the meta-learner framework, 
which decomposes the CATE estimation problem into several sub-problems,
can solve the sub-problem using any machine learning model (called a base learner),
by effectively choosing the formulation of the base learners and loss functions,
we can obtain the optimal task-specific parameters for each sub-problem,
making it possible to obtain the optimal task-specific parameters for the overall problem in a closed form as well.
We detail this effective choice of the base learners and loss functions in Section~\ref{sec:solver}.

In what follows, we illustrate the general procedure for task adaptation
by following a meta-learner method called 
the DR-Learner~\cite{kennedy2020optimal},
which has achieved successful CATE estimation performance~\cite{curth2021nonparametric}.
Note that in our framework, we can use other meta-learner methods,
such as the regression adjustment learner (RA-Learner)
or plugin learners~\cite{curth2021nonparametric},
as detailed in Section~\ref{sec:plugin}.

The DR-Learner decomposes the CATE estimation task into three sub-problems,
i.e., the estimation of propensity score model $\pi$,
outcome model $\mu_a$ for each treatment assignment $a\in\{0,1\}$,
and pseudo outcome model $\tau$.
It estimates the CATE in two steps.
In the first step,
it learns the propensity score model and the outcome models
We formulate these models as follows:
\begin{align}
  \pi(\vec{x};\bm{\theta}_{\rm{p}}^{(t)},\bm{\phi}_{\rm{p}})\approx \mathrm{P}^{(t)}(A=1|\vec{X}=\vec{x}),
  \label{eq:pi}
\end{align}
and 
\begin{align}
  \mu_{a}(\vec{x};\bm{\theta}_{a}^{(t)},\bm{\phi}_{a})\approx\mathbb{E}^{(t)}[Y|\vec{X}=\vec{x},A=a],
  \label{eq:mu}
\end{align}
where $\bm{\theta}_{\rm{p}}^{(t)}$ and $\bm{\theta}_{a}^{(t)}$ are the task-specific parameters,
and $\bm{\phi}_{\rm{p}}$ and $\bm{\phi}_{a}$ are the task-shared parameters.
The estimation targets of these models are
propensity score $\mathrm{P}^{(t)}(A=1|\vec{X}=\vec{x})$,
which is the probability of being treated when the feature vector is $\vec{x}$ in task $t$,
and conditional expected value of outcome $\mathbb{E}^{(t)}[Y|\vec{X}=\vec{x},A=a]$,
which is the average outcome in task $t$
across individuals with feature attributes $\vec{x}$ and treatment assignment $a$.
To adapt to each task $t$, we fit the task-specific parameters of the two models
by minimizing the loss on support set $\mathcal{S}$
while fixing the task-shared parameters:
\begin{align}
  \hat{\bm{\theta}}_{\rm{p}}^{(t)}=\argmin_{\bm{\theta}_{\rm{p}}}\sum_{(\vec{x}^{\rm{s}},a^{\rm{s}})\in\mathcal{S}}\ell_{\rm{p}}\left(a^{\rm{s}},\pi(\vec{x}^{\rm{s}};\bm{\theta}_{\rm{p}},\bm{\phi}_{\rm{p}})\right),
  \label{eq:hat_theta_pi}
\end{align}
\begin{align}
  \hat{\bm{\theta}}_{a}^{(t)}=\argmin_{\bm{\theta}_{a}}\sum_{(\vec{x}^{\rm{s}},y^{\rm{s}})\in\mathcal{S}_{a}}\ell_{\rm{a}}\left(y^{\rm{s}},\mu_{a}(\vec{x}^{\rm{s}};\bm{\theta}_{a},\bm{\phi}_{a})\right),
  \label{eq:hat_theta_mu}  
\end{align}
where $\ell_{\rm{p}}$ and $\ell_{\rm{a}}$ are loss functions,
and
\begin{align}
  \mathcal{S}_{a}=\{(\vec{x}^{\rm{s}},y^{\rm{s}})|(\vec{x}^{\rm{s}},y^{\rm{s}},a^{\rm{s}})\in\mathcal{S},a^{\rm{s}}=a\},
\end{align}
is a set of support instances with treatment assignment $a$,
which is assumed to be a non-empty set for all $a\in\{0,1\}$.

In the second step,
the DR-Learner fits pseudo outcome model $\tau$,
which takes as input feature vector $\vec{X}$ and outputs a CATE estimate 
as a representation of outcome, called pseudo outcome $\tilde{Y}$.
In our setup, the pseudo outcome regression model corresponds to 
CATE estimation model $\tau$ in Eq.~(\ref{eq:tau}),
which contains task-specific parameters $\bm{\theta}_{\rm{y}}^{(t)}$ and task-shared parameters $\bm{\phi}_{\rm{y}}$.
To fit task-specific parameters $\bm{\theta}_{\rm{y}}^{(t)}$,
we perform pseudo outcome regression as follows.
Using the optimal task-specific parameter values of the propensity score model and outcome models, 
i.e., $\hat{\bm{\theta}}_{\rm{p}}^{(t)}$, $\hat{\bm{\theta}}_{0}^{(t)}$, and $\hat{\bm{\theta}}_{1}^{(t)}$
in Eqs.~(\ref{eq:hat_theta_pi}, \ref{eq:hat_theta_mu}),
we first compute the pseudo outcome for each support instance $n\in\{1,\dots,N^\mathrm{s}\}$ by
\begin{align}
  \tilde{y}^{\rm{s}}_{n}&=
  \left(\frac{a^{\rm{s}}_{n}}{\pi(\vec{x}^{\rm{s}}_{n};\hat{\bm{\theta}}_{\rm{p}}^{(t)},\bm{\phi}_{\rm{p}})}-
  \frac{1-a^{\rm{s}}_{n}}{1-\pi(\vec{x}^{\rm{s}}_{n};\hat{\bm{\theta}}_{\rm{p}}^{(t)},\bm{\phi}_{\rm{p}})}\right)y^{\rm{s}}_{n}
  \nonumber\\
  &+
  \Bigl[\Bigl(1-\frac{a^{\rm{s}}_{n}}{\pi(\vec{x}^{\rm{s}}_{n};\hat{\bm{\theta}}_{\rm{p}}^{(t)},\bm{\phi}_{\rm{p}})}\Bigr)\mu_{1}(\vec{x}^{\rm{s}}_{n};\hat{\bm{\theta}}_{1}^{(t)},\bm{\phi}_{1})
    -\Bigl(1-\frac{1-a^{\rm{s}}_{n}}{1-\pi(\vec{x}^{\rm{s}}_{n};\hat{\bm{\theta}}_{\rm{p}}^{(t)},\bm{\phi}_{\rm{p}})}\Bigr)
    \mu_{0}(\vec{x}^{\rm{s}}_{n};\hat{\bm{\theta}}_{0}^{(t)},\bm{\phi}_{0})\Bigr].
  \label{eq:tilde_y}
\end{align}
Then we employ these pseudo outcome values
to construct paired data 
$\{(\vec{x}^{\mathrm{s}}_{n},\tilde{y}^{\mathrm{s}}_{n})\}_{n=1}^{N^{\mathrm{s}}}$.
By minimizing the regression loss on these paired data,
we fit task-specific parameters $\bm{\theta}^{(t)}_{\mathrm{y}}$ in $\tau$ as follows:
\begin{align}
  \hat{\bm{\theta}}_{\rm{y}}^{(t)}=\argmin_{\bm{\theta}_{\rm{y}}}\sum_{(\vec{x}^{\rm{s}},\tilde{y}^{\rm{s}})\in\mathcal{S}}\ell_{\rm{y}}\left(\tilde{y}^{\rm{s}},\tau(\vec{x}^{\rm{s}};\bm{\theta}_{\rm{y}},\bm{\phi}_{\rm{y}})\right),
  \label{eq:hat_theta_tau}  
\end{align}
According to~\cite{kennedy2020optimal},
such a pseudo outcome regression procedure yields an unbiased CATE estimator
(in our setup, $\mathbb{E}^{(t)}[\tilde{Y}|\vec{X}=\vec{x}]=\mathbb{E}^{(t)}[Y(1)-Y(0)|\vec{X}=\vec{x}]$)
if either the propensity score or the outcome models are correctly specified.

Note that in Eqs.~(\ref{eq:hat_theta_pi}, \ref{eq:hat_theta_mu}, \ref{eq:hat_theta_tau}),
unlike the original DR-Learner~\cite{kennedy2020optimal},
we fit the three models to the same data in support set $\mathcal{S}$ without data splitting.
This is because data sample size $N^{\mathrm{s}}$ is very small, and in such a low sample-size setting, as shown by~\cite{curth2021nonparametric}, the DR-Learner without data splitting works better in practice.

Although in Eqs.~(\ref{eq:hat_theta_pi}, \ref{eq:hat_theta_mu}, \ref{eq:hat_theta_tau}),
we can use any machine learning models for $\pi$, $\mu_{a}$, and $\tau$,
and any loss functions for $\ell_{\rm{p}}$, $\ell_{\rm{a}}$, and $\ell_{\rm{y}}$,
depending on their choices, the minimization problems
in Eqs.~(\ref{eq:hat_theta_pi}, \ref{eq:hat_theta_mu}, \ref{eq:hat_theta_tau})
can be computationally expensive.
For instance,
if we formulate propensity score $\pi$ with the widely-used logistic regression model,
we cannot analytically obtain the optimal task-specific parameters in a closed form;
hence, we cannot avoid performing a computationally demanding gradient-based iterative optimization.
To resolve this issue, 
we develop a differentiable closed-form solver for task adaptation
by adopting effective formulation for the meta-learner methods
as described in Section~\ref{sec:solver}.

\subsection{Differentiable closed-form solver for task adaptation}
\label{sec:solver}

To analytically obtain the optimal task-specific parameters, 
we present an effective choice of the models and the loss functions based on recently proposed meta-learning methods.

We formulate propensity score model $\pi$
using the prototypical network~\cite{snell2017prototypical},
which is a widely-used classification model in meta-learning:
\begin{align}
  \pi(\vec{x};\bm{\theta}_{\rm{p}},\bm{\phi}_{\rm{p}})
  =\frac{\exp(-\| f_{{\rm p}}(\vec{x};\bm{\phi}_{{\rm p}})-\bm{\theta}_{{\rm p}1}\|^{2})}{\sum_{a'=0}^{1}\exp(-\| f_{{\rm p}}(\vec{x};\bm{\phi}_{{\rm p}})-\bm{\theta}_{{\rm p}a'}\|^{2})},
  \label{eq:proto}
\end{align}
where $f_{{\rm p}}(\vec{x};\bm{\phi}_{{\rm p}}):\mathcal{X}\rightarrow\mathbb{R}^{K_{{\rm p}}}$
is a task-shared neural-network-based encoder,
$\bm{\theta}_{{\rm p}a}\in\mathbb{R}^{K_{{\rm p}}}$
is the mean vector of the task-specific Gaussian distribution
for treatment assignment $a$,
and $\bm{\theta}_{{\rm p}}=\{\bm{\theta}_{{\rm p}0},\bm{\theta}_{{\rm p}1}\}.$
The propensity score is estimated based on the distance to the mean vectors in the encoded space.
When using the negative log-likelihood as loss $\ell_{{\rm p}}$,
we can analytically compute the optimal task-specific parameters adapted to support set $\mathcal{S}$
as the empirical mean of the encoded vectors
of the support instances with treatment assignment $a\in\{0,1\}$:
\begin{align}
  \hat{\bm{\theta}}_{{\rm p}a}^{(t)}=\frac{1}{N^{\mathrm{s}}_{a}}
  \sum_{\vec{x}\in\mathcal{S}_{a}}f_{{\rm p}}(\vec{x};\bm{\phi}_{{\rm p}}),
  \label{eq:hat_theta_p}
\end{align}
where $N^{{\rm s}}_{a}=|\mathcal{S}_{a}|$ is the number of the support instance with $a$.

We formulate outcome model $\mu_{a}$ and pseudo outcome model $\tau$
using task-shared encoders with the task-specific last linear layer~\cite{bertinetto2018meta}:
\begin{align}
  \mu_{a}(\vec{x};\bm{\theta}_{a}^{(t)},\bm{\phi}_{a})
  =\bm{\theta}_{a}^{\top}f_{a}(\vec{x};\bm{\phi}_{a}),
  \label{eq:r2d2_mu}
\end{align}  
\begin{align}
  \tau(\vec{x};\bm{\theta}_{\rm{y}}^{(t)},\bm{\phi}_{\rm{y}})
  =\bm{\theta}_{\rm{y}}^{\top}f_{\rm{y}}(\vec{x};\bm{\phi}_{\rm{y}}),
  \label{eq:r2d2_tau}
\end{align}  
where $f_{a}(\vec{x};\bm{\phi}_{a}):\mathcal{X}\rightarrow\mathbb{R}^{K_{a}}$
and $f_{\rm{y}}(\vec{x};\bm{\phi}_{\rm{y}}):\mathcal{X}\rightarrow\mathbb{R}^{K_{\rm{y}}}$ are neural-network-based task-shared encoders,
and $\bm{\theta}_{a}\in\mathbb{R}^{K_{a}}$ and
$\bm{\theta}_{\rm{y}}\in\mathbb{R}^{K_{\rm{y}}}$
are task-specific linear projection vectors at the last layer.

Based on these linear models, we formulate loss functions $\ell_{\rm{a}}$ and $\ell_{\rm{y}}$
using the mean squared error plus the $\ell^{2}$-norm regularizer.
Such a formulation choice enables us to
obtain the optimal task-specific parameters adapted to support set $\mathcal{S}$
in a closed form:
\begin{align}
  \hat{\bm{\theta}}_{a}^{(t)}=
  (\vec{Z}_{a}^{\rm{s}\top}\vec{Z}_{a}^{\rm{s}}+\lambda_{a}\vec{I})^{-1}\vec{Z}_{a}^{\rm{s}}\vec{y}_{a}^{\rm{s}},
  \label{eq:hat_theta_a}
\end{align}
\begin{align}
  \hat{\bm{\theta}}_{\rm{y}}^{(t)}=
  (\vec{Z}_{\rm{y}}^{\rm{s}\top}\vec{Z}_{\rm{y}}^{\rm{s}}+\lambda_{\rm{y}}\vec{I})^{-1}\vec{Z}_{\rm{y}}^{\rm{s}}\tilde{\vec{y}}^{\rm{s}},
  \label{eq:hat_theta_y}
\end{align}
where $\vec{Z}_{a}^{\rm{s}}=(f_{a}(\vec{x}^{\rm{s}};\bm{\phi}_{a}))_{\vec{x}^{\rm{s}}\in\mathcal{S}_{a}}\in\mathbb{R}^{{N^{{\rm s}}_{a}\times K_{a}}}$
is a matrix consisting of encoded vectors of the support instances with treatment assignment $a$,
$\vec{y}_{a}^{\rm{s}}=(y^{\rm{s}})_{y^{\rm{s}}\in\mathcal{S}_{a}}\in\mathbb{R}^{N_{a}^{\mathrm{s}}}$
is its outcome vector,
$\vec{Z}_{\rm{y}}^{\rm{s}}=(f_{\rm{y}}(\vec{x}^{\rm{s}};\bm{\phi}_{\rm{y}}))_{\vec{x}^{\rm{s}}\in\mathcal{S}}\in\mathbb{R}^{{N^{\rm{s}}\times K_{\rm{y}}}}$
is a matrix consisting of encoded vectors of the support instances,
$\tilde{\vec{y}}^{\rm{s}}=(\tilde{y}^{\rm{s}})_{\tilde{y}^{\rm{s}}\in\mathcal{S}}\in\mathbb{R}^{N^{\rm{s}}}$
is the pseudo outcome vector,
$\lambda_{a}\in\mathbb{R}_{>0}$ and $\lambda_{\rm{y}}\in\mathbb{R}_{>0}$ are the non-negative scalar regularization parameters,
and $\vec{I}$ is an identity matrix.

Although assuming the linearity as shown in Eqs.~(\ref{eq:r2d2_mu}, \ref{eq:r2d2_tau})
might be restrictive in handling the complex task heterogeneity,
we can consider their nonlinear extension using Gaussian processes, as described in Section~\ref{sec:gp}.


\subsection{Learning task-shared parameters}
\label{sec:meta-learn}

So far, we have discussed how we can adapt the task-specific
parameters to each task. In this section,
we present a strategy for learning task-shared parameters to extract
common knowledge across multiple tasks.
The task-shared parameters in our model
are parameters in task-shared neural-network-based encoders
and regularization parameters,
i.e., $\bm{\Phi}=\{\bm{\phi}_{\rm{p}},\bm{\phi}_{0},\bm{\phi}_{1},\bm{\phi}_{\rm{y}},\lambda_{0},\lambda_{1},\lambda_{\rm{y}}\}$.

We learn these parameters such that
the expected CATE estimation performance is improved.
In usual meta-learning setups,
we can improve such a performance by minimizing the difference between the true labels and the estimated ones
over the meta-training tasks.
However, in our setup,
we cannot obtain the true CATE values because we can never jointly observe two potential outcomes: $Y(0)$ and $Y(1)$.

To overcome this difficulty, as alternatives to the true CATE values,
we utilize pseudo CATEs, which
are estimated with a CATE estimation model fitted to
large observational data.
Such alternatives will be reliable
as the data sample size increases for the CATE model fitting.
We prepare such pseudo CATEs at the beginning of
the meta-training phase by taking the following steps for each task $t$.
First, we fit a CATE model to data $\mathcal{D}_{t}$;
here we can use any existing estimation model.
Then we estimate the CATEs with the fitted model
to obtain pseudo CATEs
$\{\tilde{\tau}_{tn}\}_{n=1}^{N_{t}}$,
where $\tilde{\tau}_{tn}$ denotes the pseudo CATE of the $n$th instance in $\mathcal{D}_{t}$.
In our experiments, we estimated pseudo CATEs
using the RA-Learner~\cite{curth2021nonparametric}.

By minimizing the expected difference
between these estimated pseudo CATEs and
the CATEs estimated by our task-adapted CATE model with small data,
we estimate task-shared parameters $\bm{\Phi}$ as follows:
\begin{align}
  \hat{\bm{\Phi}}=
  \argmin_{\bm{\Phi}} \mathbb{E}_{t}\left[\mathbb{E}_{(\mathcal{S},\mathcal{Q})\sim\mathcal{D}_{t}}\left[
      \ell(\mathcal{S},\mathcal{Q},\bm{\Phi})
      \right]
    \right],
  \label{eq:hat_phi}  
\end{align}
where $\ell$ is the following CATE loss function:
\begin{align}
  \ell(\mathcal{S},\mathcal{Q},\bm{\Phi})
  =
      \sum_{(\vec{x}_{n}^{\rm{q}},\tilde{\tau}_{n}^{\rm{q}})\in\mathcal{Q}}
      \parallel\tilde{\tau}_{n}^{\rm{q}}-\tau(\vec{x}_{n}^{\mathrm{q}};\hat{\bm{\theta}}_{\rm{y}}^{(t)},\bm{\Phi})\parallel^{2},
      \label{eq:L}
\end{align}
where $\mathbb{E}_{t}$ is the expectation over tasks,
$\mathbb{E}_{(\mathcal{S},\mathcal{Q})\sim\mathcal{D}_{t}}$ is the expectation over
support and query sets from data in task $t$,
and $\hat{\bm{\theta}}_{\rm{y}}^{(t)}$ is the adapted task-specific parameters
given by Eq.~(\ref{eq:hat_theta_tau}) or Eq.~(\ref{eq:hat_theta_y}).

In our CATE loss function in Eq.~(\ref{eq:L}), 
the estimated difference is evaluated over a data sample subset (called a query set),
denoted by $\mathcal{Q}=\{(\vec{x}_{n}^{\rm{q}},\tilde{\tau}^{\rm{q}}_{n})\}_{n=1}^{N^{\rm{q}}}$,
which consists of feature vectors and their pseudo CATEs.
Using a query set that does not overlap with the support set,
we can evaluate and maximize the test CATE estimation performance.
Note that unlike Eq.~(\ref{eq:tau}),
the input of $\tau$ in Eq.~(\ref{eq:L})
include all task-shared parameters $\bm{\Phi}$.
This is because with our formulation,
CATE estimator $\tau$ depends on
all task-shared parameters $\bm{\Phi}$
through
Eqs.~(\ref{eq:tilde_y}, \ref{eq:proto}, \ref{eq:r2d2_mu}, \ref{eq:r2d2_tau}).

Our objective function in Eq.~(\ref{eq:hat_phi}) is the expected value of the CATE loss function $\ell$ over tasks.
As with the standard meta-learning methods~\cite{finn2017model},
by minimizing such an expected loss,
we can expect that the test performance increases as the number of meta-training tasks increases
under the assumption that the task distributions are identical between the meta-training and meta-test phases.
Optimizing our objective function
is to solve a tri-level optimization problem,
where the outer optimization estimates task-shared parameters
by minimizing the CATE loss in Eq.~(\ref{eq:hat_phi}),
the middle optimization adapts task-specific parameters of CATE estimation model $\tau$
by minimizing the pseudo outcome loss in Eq.~(\ref{eq:hat_theta_tau}),
and the inner optimization adapts task-specific parameters of
propensity score $\pi$ and outcome model $\mu_{a}$ 
by minimizing the classification and regression losses
in Eqs.~(\ref{eq:hat_theta_pi}, \ref{eq:hat_theta_mu}).
Since the solutions to the middle and inner optimizations
are given in differentiable closed forms, as shown in
Eqs.~(\ref{eq:hat_theta_p}, \ref{eq:hat_theta_a}, \ref{eq:hat_theta_y}),
we can solve the tri-level optimization problem as
a single-level optimization,
which can be efficiently solved with a stochastic gradient descent method.

Algorithm~\ref{alg:train} shows our proposed meta-training procedure for CATE estimation.
The expectation in Eq.~(\ref{eq:hat_phi}) is approximated
with the Monte Carlo method,
where tasks, support sets, and query sets are randomly sampled from
the given meta-training data.
We sample the support and query sets for each treatment assignment.
The time complexity of computing the support-adapted parameters
of the propensity score model is $O(N^{\rm{s}})$ in Eq.~(\ref{eq:hat_theta_p}).
Those of the outcome and pseudo outcome models
are $O(K_{a}^{3})$ and $O(K_{\rm{y}}^{3})$ due to the matrix inverse in Eqs.~(\ref{eq:hat_theta_a}, \ref{eq:hat_theta_y}),
which can be
$O(N^{{\rm s}3}_{a})$ and $O(N^{{\rm s}3})$ using the Woodbury formula~\cite{bertinetto2018meta}.

\begin{algorithm}[t!]
  \centering
  \caption{Meta-learning procedure of our model.}
  \label{alg:train}
  \begin{algorithmic}[1]
    \renewcommand{\algorithmicrequire}{\textbf{Input:}}
    \renewcommand{\algorithmicensure}{\textbf{Output:}}
    \REQUIRE{Meta-training data $\{\mathcal{D}_{t}\}_{t=1}^{T}$,
      number of untreated support instances $N^{\mathrm{s}}_{0}$,
      number of treated support instances $N^{\mathrm{s}}_{1}$,      
      number of untreated query instances $N^{\mathrm{q}}_{0}$,
      number of treated query instances $N^{\mathrm{q}}_{1}$.}
    \ENSURE{Trained task-shared parameters $\bm{\Phi}$.}
    \FOR{each task $t=1,\cdots,T$}
    \STATE Estimate pseudo CATEs $\{\tilde{\tau}_{tn}\}_{n=1}^{N_{t}}$
    using large data $\mathcal{D}_{t}$.
    \ENDFOR
    \STATE Initialize task-shared parameters $\bm{\Phi}$.
    \WHILE{End condition is satisfied}
    \STATE Randomly select task $t$ from $\{1,\cdots,T\}$.
    \FOR{each treatment assignment $a=0,1$}
    \STATE Randomly sample $N^{\mathrm{s}}_{a}$ instances
    with treatment assignment $a$
    for support set $\mathcal{S}_{a}$ from $\mathcal{D}_{t}$.
    \STATE Randomly sample $N^{\mathrm{q}}_{a}$ instances
    with treatment assignment $a$
    for query set $\mathcal{Q}_{a}$ from $\mathcal{D}_{t}\setminus\mathcal{S}_{a}$.    
    \ENDFOR
    \STATE Construct support and query sets $\mathcal{S}=\mathcal{S}_{0}\cup\mathcal{S}_{1}$, $\mathcal{Q}=\mathcal{Q}_{0}\cup\mathcal{Q}_{1}$.
    \STATE Compute the task-specific parameters for the propensity score and outcome models by Eqs.~(\ref{eq:hat_theta_p}, \ref{eq:hat_theta_a}).
    \STATE Compute the pseudo outcomes for each instance in support set $\mathcal{S}$ by Eq.~(\ref{eq:tilde_y}).
    \STATE Compute the task-specific parameters for the pseudo outcome model by Eq.~(\ref{eq:hat_theta_y}).
    \STATE Compute CATE loss $\ell$ in Eq.~(\ref{eq:L}).
    \STATE Update task-shared parameters $\bm{\Phi}$ using a stochastic gradient method.
    \ENDWHILE
  \end{algorithmic}
\end{algorithm}


\subsection{Variants}
\label{sec:variant}

\subsubsection{CATE estimation with other meta-learners}
\label{sec:plugin}
  
Whereas we have explained how we perform meta-learning for CATE estimation with the DR-Learner, 
we can also use other meta-learner methods instead of the DR-Learner.
In this section, we take two examples,
the RA-Learner and plugin learners~\cite{curth2021nonparametric}.

The RA-Learner decomposes the CATE estimation problem into
the two sub-problems, each of which aims to estimate
the outcome for each treatment assignment and the pseudo outcome, respectively.
A key difference between the DR-Learner and the RA-Learner
is the formulation of the pseudo outcome;
in our setup, the pseudo outcome of the RA-Learner is constructed by
\begin{align}
\tilde{y}_{\mathrm{RA}n}^{\rm{s}}=
  a^{\rm{s}}_{n}\left(y^{\rm{s}}_{n}-\mu_{0}(\vec{x}^{\rm{s}}_{n};\hat{\bm{\theta}}_{0}^{(t)},\bm{\phi}_{0})\right)
  +(1-a^{\rm{s}}_{n})\left(\mu_{1}(\vec{x}^{\rm{s}}_{n};\hat{\bm{\theta}}_{1}^{(t)},\bm{\phi}_{1})-y^{\rm{s}}_{n}\right),
  \label{eq:ytilde_ra}
\end{align}
where $\mu_{a}(\vec{x}^{\rm{s}}_{n};\hat{\bm{\theta}}_{a}^{(t)},\bm{\phi}_{a})$ is
given by Eqs.~(\ref{eq:r2d2_mu},\ref{eq:hat_theta_a}).
Using Eq.~(\ref{eq:ytilde_ra}) instead of Eq.~(\ref{eq:tilde_y})
yields the proposed method with the RA-Learner.
Regardless of such a formulation difference in the pseudo outcome, 
our closed-form solvers for task adaptation can be formulated in the same way.
Hence, as with the DR-Learner,
we can obtain the optimal task-specific parameters in a closed form for the RA-Learner.

The plugin learner decomposes the CATE estimation task into the estimation tasks of
the outcome for each treatment assignment.
With the plugin learner, the CATE is estimated by the difference in outcome models:
\begin{align}
  \tau(\vec{x};\hat{\bm{\theta}}_{\mathrm{y}}^{(t)},\bm{\phi}_{\mathrm{y}})
    =\mu_{1}(\vec{x};\hat{\bm{\theta}}_{1}^{(t)},\bm{\phi}_{1})-\mu_{0}(\vec{x};\hat{\bm{\theta}}_{0}^{(t)},\bm{\phi}_{0}),
  \label{eq:tau_plugin}
\end{align}
where $\hat{\bm{\theta}}_{\mathrm{y}}^{(t)}=\{\hat{\bm{\theta}}_{0}^{(t)},\hat{\bm{\theta}}_{1}^{(t)}\}$
and $\bm{\phi}_{\mathrm{y}}=\{\bm{\phi}_{0},\bm{\phi}_{1}\}$.
Using Eq.~(\ref{eq:tau_plugin}) instead of Eq.~(\ref{eq:r2d2_tau})
yields the proposed method with the plugin learner.
As with the DR-Learner,
we can obtain the optimal task-specific parameter values of
these outcome models in a closed form.

\subsubsection{Gaussian process-based task adaptation}
\label{sec:gp}

In Section~\ref{sec:solver}, we use
task-shared neural-network-based encoders with task-specific linear models at the last layer
for the outcome and pseudo outcome models.
Instead of the linear models in Eqs.~(\ref{eq:r2d2_mu},\ref{eq:r2d2_tau}),
we can use Gaussian process models at the last layer,
which also gives closed-form solutions for task adaptation.
Formally, as with the existing meta-learning method~\cite{patacchiola2020bayesian},
for given input feature vector $\vec{x}$,
outcome model $\mu_{a}$ and pseudo outcome model $\tau$ are assumed to be generated from Gaussian processes as follows:
\begin{align}
  \mu_{a}(\vec{x};\bm{\theta}_{a}^{(t)},\bm{\phi}_{a})
  \sim\mathrm{GP}(\vec{0},k_{a}(f_{a}(\vec{x};\bm{\phi}_{a}),f_{a}(\vec{x}';\bm{\phi}_{a}))),
  \label{eq:gp_prior_mu}
\end{align}
\begin{align}
  \tau(\vec{x};\bm{\theta}_{\mathrm{y}}^{(t)},\bm{\phi}_{\mathrm{y}})
  \sim\mathrm{GP}(\vec{0},k_{\mathrm{y}}(f_{\mathrm{y}}(\vec{x};\bm{\phi}_{\mathrm{y}}),f_{\mathrm{y}}(\vec{x}';\bm{\phi}_{\mathrm{y}}))),
  \label{eq:gp_prior_tau}
\end{align}
where $\mathrm{GP}(\vec{0},k(\vec{x},\vec{x}'))$ is the Gaussian process with the zero mean function and kernel function $k$,
and $k_{a}$ and $k_{\mathrm{y}}$ are kernel functions.
Then, the optimal task-adapted models are given in a closed form
by the posterior mean of the Gaussian processes as follows:
\begin{align}
  \mu_{a}(\vec{x};\hat{\bm{\theta}}_{a}^{(t)},\bm{\phi}_{a})
  =\vec{k}^{\mathrm{s}\top}_{a}\vec{K}^{\mathrm{s}-1}_{a}\vec{y}^{\mathrm{s}}_{a},
  \label{eq:gp_mu}
\end{align}
\begin{align}
  \tau(\vec{x};\hat{\bm{\theta}}_{\rm{y}}^{(t)},\bm{\phi}_{\rm{y}})
  =\vec{k}^{\mathrm{s}\top}_{\mathrm{y}}\vec{K}^{\mathrm{s}-1}_{\mathrm{y}}\tilde{\vec{y}}^{\mathrm{s}},  
  \label{eq:gp_tau}
\end{align}  
where $\vec{k}^{\mathrm{s}}_{a}=(k_{a}(f_{a}(\vec{x};\bm{\phi}_{a}),f_{a}(\vec{x}^{\mathrm{s}};\bm{\phi}_{a})))_{\vec{x}^{\mathrm{s}}\in\mathcal{S}_{a}}\in\mathbb{R}^{N^{\mathrm{s}}_{a}}$
is the kernel vector between $\vec{x}$ and the support instances with treatment assignment $a$,
$\vec{K}^{\mathrm{s}}_{a}=(k_{a}(f_{a}(\vec{x}^{\mathrm{s}};\bm{\phi}_{a}),f_{a}(\vec{x}^{\mathrm{s}\prime};\bm{\phi}_{a})))_{\vec{x}^{\mathrm{s}},\vec{x}^{\mathrm{s}\prime}\in\mathcal{S}_{a}}\in\mathbb{R}^{N^{\mathrm{s}}_{a}\times N^{\mathrm{s}}_{a}}$ is the kernel matrix between the support instances with treatment assignment $a$,
$\vec{k}^{\mathrm{s}}_{\mathrm{y}}=(k(f_{\mathrm{y}}(\vec{x};\bm{\phi}_{\mathrm{y}}),f_{\mathrm{y}}(\vec{x}^{\mathrm{s}};\bm{\phi}_{\mathrm{y}})))_{\vec{x}^{\mathrm{s}}\in\mathcal{S}}\in\mathbb{R}^{N^{\mathrm{s}}}$ 
is the kernel vector between $\vec{x}$ and the support instances,
and
$\vec{K}^{\mathrm{s}}_{\mathrm{y}}=(k(f_{\mathrm{y}}(\vec{x}^{\mathrm{s}};\bm{\phi}_{\mathrm{y}}),f_{\mathrm{y}}(\vec{x}^{\mathrm{s}\prime};\bm{\phi}_{\mathrm{y}})))_{\vec{x}^{\mathrm{s}},\vec{x}^{\mathrm{s}\prime}\in\mathcal{S}}\in\mathbb{R}^{N^{\mathrm{s}}\times N^{\mathrm{s}}}$ is the kernel matrix between the support instances.
Thus, thanks to the kernel trick, we can analytically obtain the optimal task-specific parameters while taking into account the complex nonlinearity in each task.

\section{Experiments}
\label{sec:experiments}

\subsection{Data}

We evaluated the proposed method using
synthetic (Synth) dataset 
and the infant health and development program (IHDP) dataset~\cite{hill2011bayesian},
which were commonly used datasets for evaluating CATE estimation.
For each dataset, we randomly split the tasks:
70\% for meta-training,
10\% for meta-validation,
and the remaining for meta-testing.

\paragraph{Synth dataset:}

We generated a dataset with 100 tasks
by following setting (ii) presented 
in~\cite{curth2021nonparametric}.
We first drew the values of 25-dimensional feature vector
$\vec{x}\in\mathbb{R}^{25}$
from the standard Gaussian distribution.
Then we sampled the values of treatment $A$ and outcome $Y$
using the three subsets of feature vector values $\vec{x}$,
denoted by $\vec{x}_{\rm{c}},\vec{x}_{\rm{o}},\vec{x}_{\tau}\in\vec{x}$.
Here features $\vec{x}_{\rm{c}}\in\mathbb{R}^{5}$ were used
as confounders that affect both treatment $A$ and outcome $Y$,
while $\vec{x}_{\rm{o}}\in\mathbb{R}^{5}$ and
$\vec{x}_{\tau}\in\mathbb{R}^{5}$ were employed as features that
influence outcome $Y$ and those that affect potential outcome $Y(1)$.
The subsets were randomly selected for each task.
The probability of the treatment assignment at task $t$ was
$\mathrm{P}^{(t)}(A=1|\vec{X}=\vec{x})=\mathrm{Sigmoid}(3 (\vec{w}_{t\mathrm{p}}^{\top}\vec{x}_{\rm{c}}^{2}-\omega))$,
where
$\vec{w}_{t{\rm p}}\in\mathbb{R}^{5}$ is the task-specific parameters,
and 
$\omega=\mathrm{Median}(\vec{w}_{t\mathrm{p}}^{\top}\vec{x}_{c}^{2})$
was used to center the propensity scores.
Baseline outcome $Y(0)$ was determined by
$\mu_{0}(\vec{x})=\vec{w}_{t0}^{\top}[\vec{x}_{\rm{c}},\vec{x}_{\rm{o}}]$,
where
$\vec{w}_{t0}\in\mathbb{R}^{10}$ is the task-specific parameters,
and
$[\cdot,\cdot]$ is the concatenation of vectors.
Treated outcome $Y(1)$ was determined by
$\mu_{1}(\vec{x})=\mu_{0}(\vec{x})+\vec{w}_{t1}^{\top}\vec{x}_{\tau}$,
where $\vec{w}_{t1}\in\mathbb{R}^{5}$ is the task-specific parameters.
Outcome $Y=(1-A)Y(0)+AY(1)$ was determined
from the potential outcomes with standard Gaussian noise.
We used task-specific parameters $\vec{w}_{t\rm{p}}, \vec{w}_{t0}, \vec{w}_{t1}$
to make the data heterogeneous across tasks.

\paragraph{IHDP dataset:} The Infant Health and Development Program (IHDP) dataset~\cite{hill2011bayesian}
is a popular benchmark dataset related to a
randomized experiment that assesses the heterogeneous effects
of an education program on the future cognitive scores across infants.
It contains real-world records of 747 infants (139 of whom are treated
and 608 are untreated), including 25 features about them.
The outcome values were simulated
based on setting B described in~\cite{hill2011bayesian}.
Since ground truth causal effects are never observed in real-world datasets due to their counterfactuality,
semi-synthetic data are needed to be able to evaluate methods for estimating CATE.
We used the data on 100 simulations provided by~\cite{johansson2016learning} 
as the data of 100 tasks.

\subsection{Compared methods}

We compared the proposed method
(Ours, which is equipped with the models presented in Section~\ref{sec:solver})
with the following methods:
two variants of meta-learning methods with DR-Learners~\cite{kennedy2020optimal} (DR-CFS and DR-ML),
a meta-learning method with the CFR method
(Meta-CI)~\cite{shalit2017estimating},
multi-task learning with the meta-training data (MT),
T-Learner (TL),
S-Learner (SL),
X-Learner (XL)~\cite{kunzel2019metalearners},
DR-Learner (DRL),
causal forest (CF)~\cite{wager2018estimation},
and the difference in the mean of outcomes (Mean).
Here, Ours, DR-CFS, DR-ML, and Meta-CI perform meta-learning;
DR-ML and Meta-CI employ model-agnostic meta-learning~\cite{finn2017model},
Ours and DR-CFS are founded on the model choice that yields closed-form solvers
for task adaptation~\cite{bertinetto2018meta,snell2017prototypical}.
Among these four methods, only DR-CFS does not employ pseudo CATEs but
minimizes the regression and classification losses.
Although MT performs multi-task learning, the other methods (i.e., TL, SL, XL, DRL, CF, and Mean)
are trained for each task.

In DR-CFS,
the DR-Learner was used, as in the proposed method,
while the task-shared parameters were meta-learned by minimizing the test classification loss of the propensity score model and the test regression losses of the outcome and pseudo outcome models, as in the existing meta-learning methods.
In DR-ML, the DR-Learner was modeled by neural networks,
where meta-learning was performed by model-agnostic meta-learning (MAML)~\cite{finn2017model}
by minimizing the CATE loss using pseudo CATEs.
MAML is a widely-used meta-learning method 
that performs task adaptation by iterative gradient descent steps.
In Meta-CI,
we employed the CFR method~\cite{shalit2017estimating} with
maximum mean discrepancy (MMD) and performed
meta-learning based on MAML that minimizes
the CATE loss using pseudo CATEs.
As regards MT, we formulated the
estimation models of the DR-Learner by
neural-network-based encoders shared across all tasks and task-specific linear final layers, as with the proposed method,
and trained their parameters
by minimizing the training classification and regression
losses on the meta-training data.

\subsection{Settings}

As the encoders for the proposed method, DR-CFS, Meta-CI, and MT,
we used three-layered feed-forward neural networks with 32 hidden and output units,
where we shared the parameters between $f_{0}$ and $f_{1}$.
For the neural networks in DR-ML,
we used four-layered feed-forward neural networks with 32 hidden units.
In Meta-CI, treated and untreated outcomes were modeled by
three-layered feed-forward neural networks with 32 hidden output units that took
encoded vectors as input,
and the MMD was computed with the RBF kernel,
where the kernel parameters were meta-trained.
In the MAML-based methods (i.e., DR-ML and Meta-CI),
we used five inner gradient descent epochs.
In neural-network-based methods, i.e., Ours, DR-CFS, DR-ML, Meta-CI, and MT, 
we used the rectified linear unit (ReLU) as the activation function,
Adam~\cite{kingma2014adam} with learning rate $10^{-3}$,
and a batch size of 32 tasks.
The maximum number of meta-training epochs was 5,000, and
the meta-validation data were used for early stopping.
We estimated pseudo CATEs
using the RA-Learner~\cite{curth2021nonparametric}
with neural-network-based base learners.
We implemented the neural-network-based methods with PyTorch~\cite{paszke2019pytorch}.
For the implementations of SL, TL, XL, DRL, and CF,
we used EconML~\cite{econml}, which is a Python package for estimating heterogeneous treatment effects,
where we used its default parameter settings and linear regression as base learners.

\subsection{Results}
\label{sec:results}

As a performance measure,
we used the mean squared error between the true and estimated
treatment effect values,
which is called the {\it precision in estimation of heterogeneous effect} (PEHE).
We averaged PEHE on the meta-test data
over 30 experiments with different meta-training, validation, and test data splits.
In particular, we evaluated each of the three cases,
where the number of the support instances was $N^{\rm{s}}=\{6,10,14\}$, and
the number of query instances was $N^{\rm{q}}=40$.
Here, in each support and query set, the number of untreated instances was set to the number of treated instances,
i.e., $N^{\mathrm{s}}_{0}=N^{\mathrm{s}}_{1}$, and $N^{\mathrm{q}}_{0}=N^{\mathrm{q}}_{1}$.

Table~\ref{tab:mse} presents the average PEHE of each method.
Our proposed method achieved the best performance in every case.
The PEHE of DR-CFS was higher than our proposed method, although
both methods used the same model architecture.
This is because DR-CFS performed meta-learning by minimizing the regression
and classification losses. On the other hand,
the proposed method performs meta-learning by maximizing the CATE estimation
performance using pseudo CATEs.
Although DR-ML and Meta-CI also use pseudo CATEs
since they did not obtain the optimal task-specific parameters
in a closed form, their PEHEs were higher than our method.
This result indicates the effectiveness of
our strategy that optimizes the task-specific parameters with closed-form solvers.

We confirmed the effectiveness of our meta-learning approach
by comparing it with the methods based on multi-task learning (i.e., MT)
and those for a single CATE estimation task (i.e., TL, SL, XL, DRL, CF, and Mean).
The former baseline yielded high PEHE because multi-task learning
is unsuitable for a few-shot setting.
The latter baselines worked poorly because they could not
utilize the data in different tasks. Among these baselines,
DRL yielded exceedingly high PEHEs because it suffered from
the estimation instability of inverse propensity scores due to data scarcity.
Although our proposed method also infers the inverse propensity scores,
since it learns the propensity score model by meta-learning that improves the estimation performance on a few data,
it robustly estimated the inverse propensity scores,
which resulted in its high CATE estimation performance.

\begin{figure}[t!]
  \centering
  \begin{tabular}{cc}
  Synth & IHDP \\
  \includegraphics[width=16em]{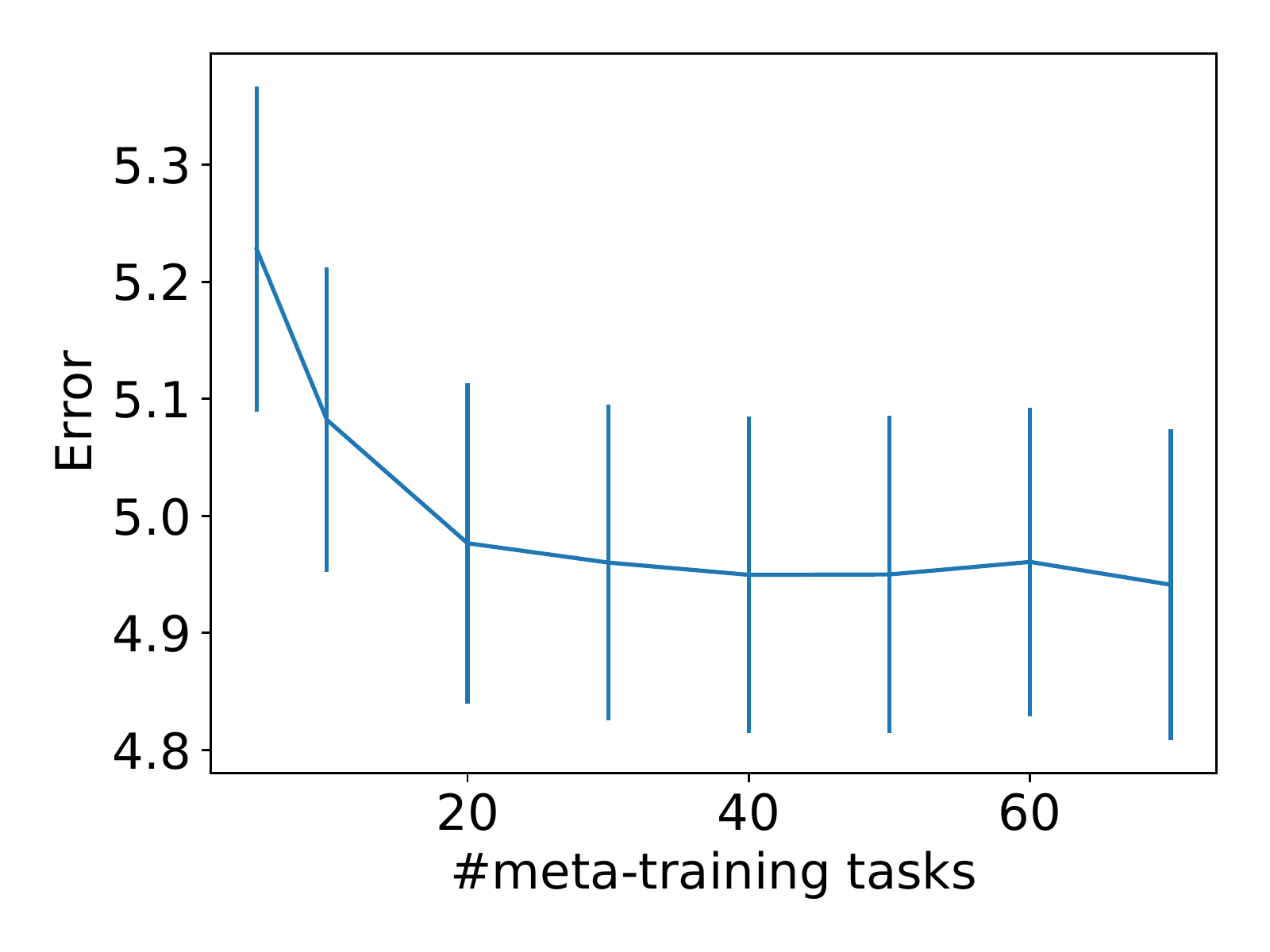}&
  \includegraphics[width=16em]{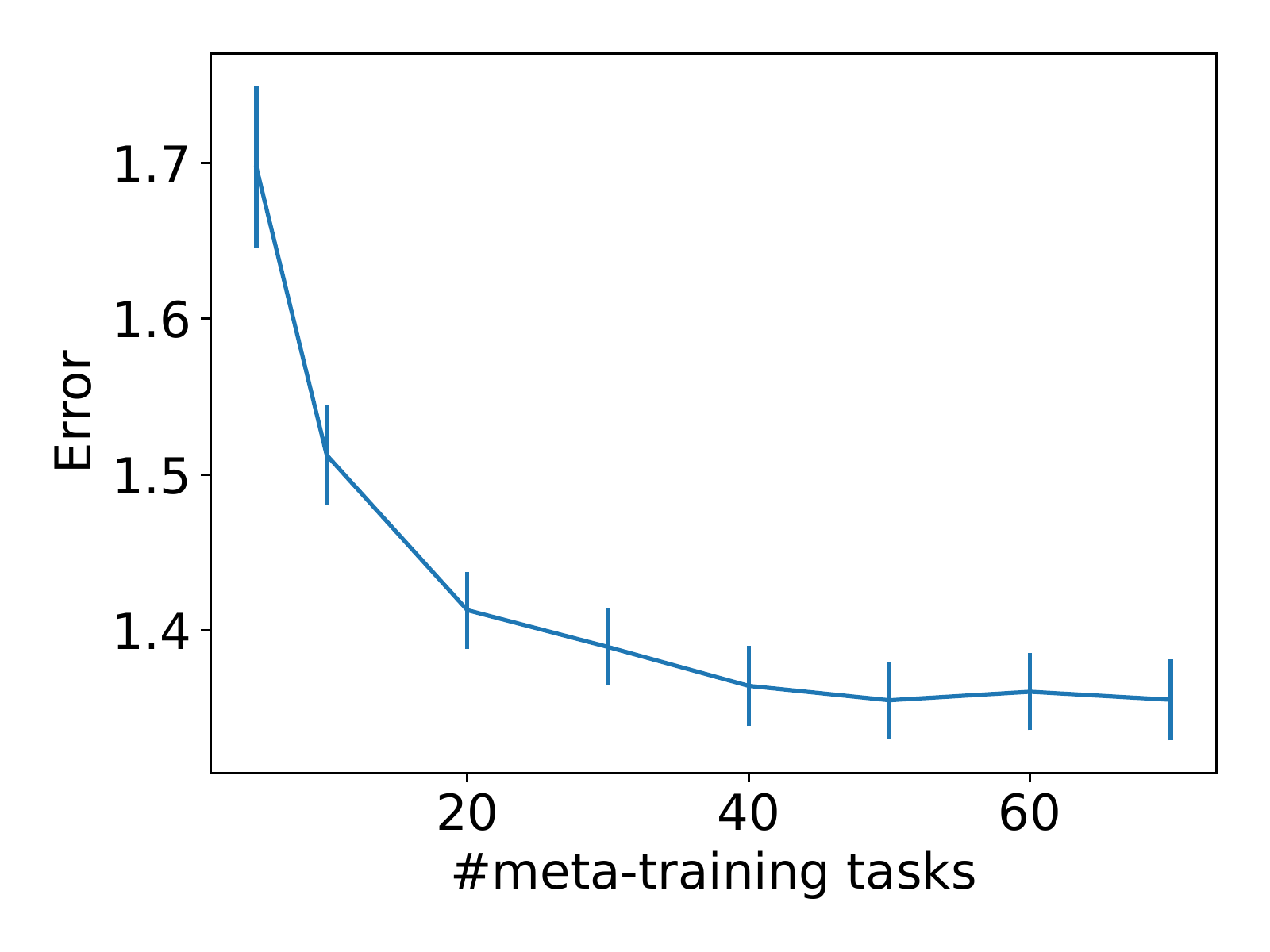}\\
  \end{tabular}
  \caption{Average PEHE and its standard error by proposed method with different numbers of meta-training tasks with $N^{\rm{s}}=6$.}
  \label{fig:n_dataset}
\end{figure}

\begin{figure}[t!]
  \centering
  \begin{tabular}{cc}
  Synth & IHDP \\
  \includegraphics[width=16em]{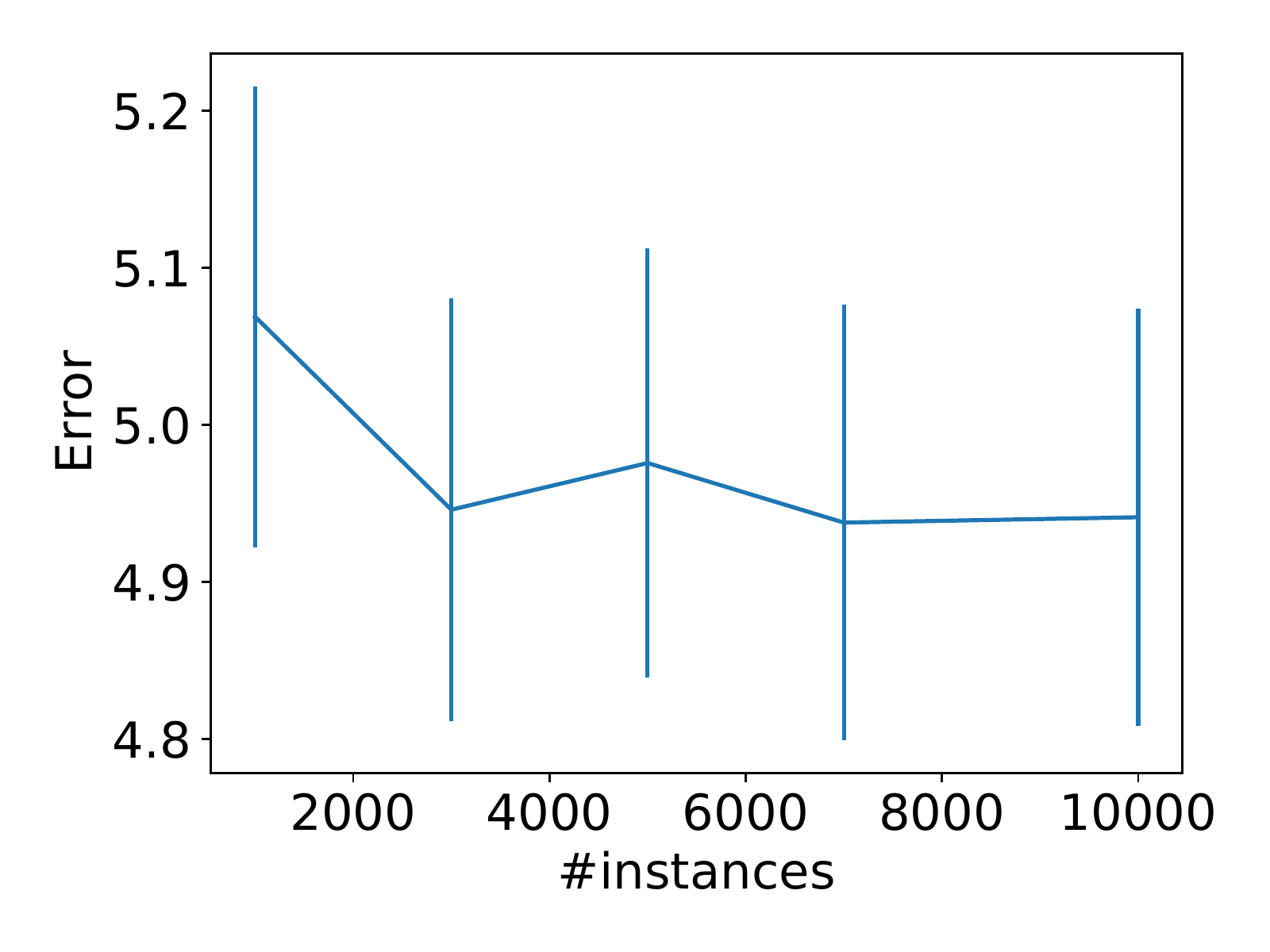}&
  \includegraphics[width=16em]{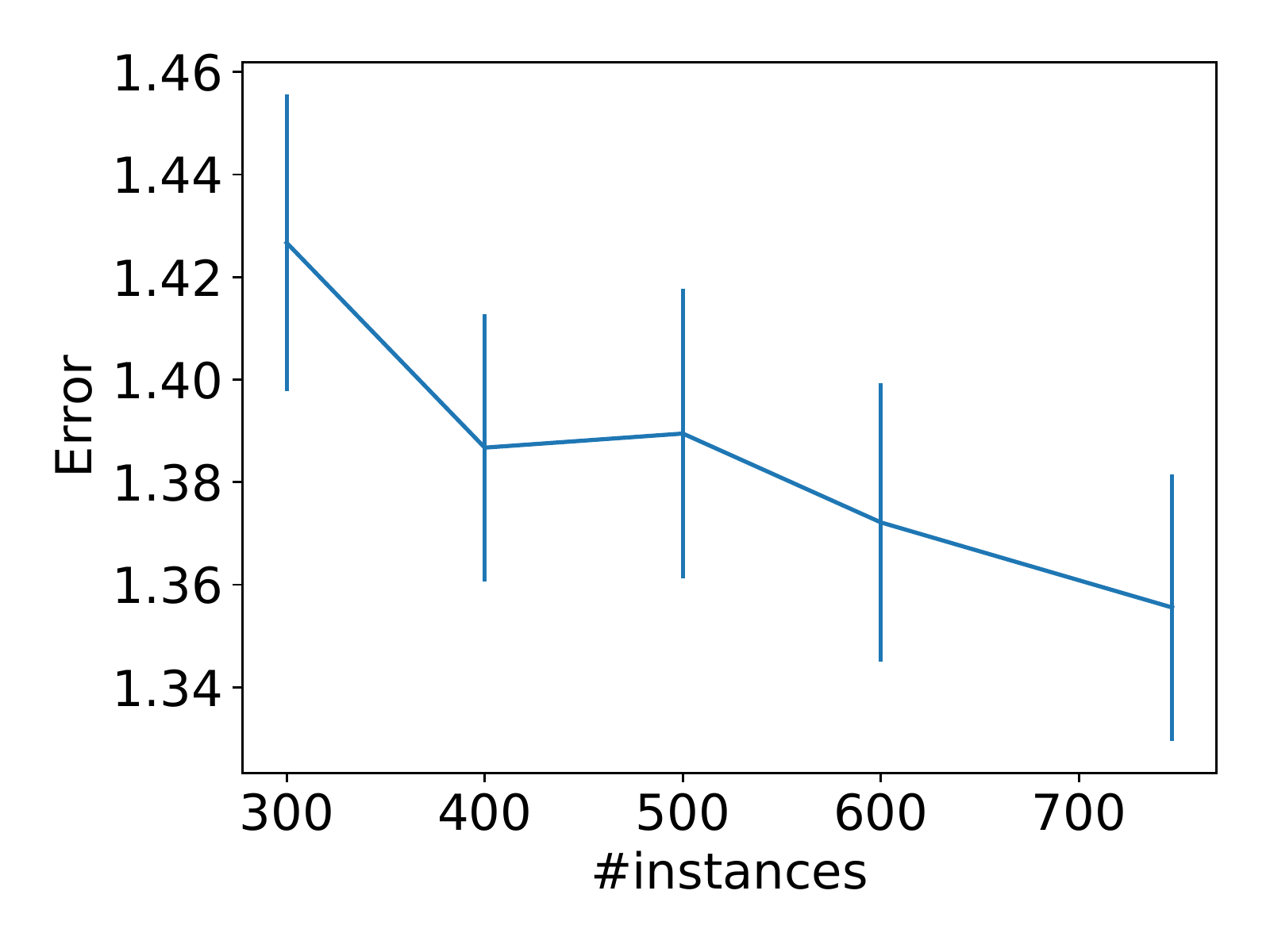}\\
  \end{tabular}
  \caption{Average PEHE and its standard error by proposed method with different numbers of instances per tasks with $N^{\rm{s}}=6$.}
  \label{fig:n_sample}
\end{figure}

We investigated how greatly the number of meta-training tasks influenced the
performance of our method.
Figure~\ref{fig:n_dataset} illustrates the PEHEs of the Synth
and IHDP datasets when the number of meta-training tasks increases.
These results indicate that performing meta-learning with
a large number of tasks is critical to improve the CATE estimation performance.
In addition, we examined how strongly the data sample size of a meta-training task affects
our method's performance.
Figure~\ref{fig:n_sample} displays the results.
As expected, as the sample size increases,
the CATE estimation performance improved.

We evaluated the performance of our method when using other CATE estimation methods
than the DR-Learner. In particular, we tested
the RA-Learner (RA)~\cite{curth2021nonparametric}
and plugin learners (Plugin) using the same model architectures as the DR-Learner.
As described in Section~\ref{sec:plugin}, with all these CATE estimation methods,
our proposed method offers differentiable closed-form solvers for task adaptation.
Table~\ref{tab:variant} shows the results.
Although all the learners were comparable on
the Synth dataset with $N^{\rm{s}}=6,10$
and IHDP dataset with $N^{\rm{s}}=14$,
the DR-Learner achieved better performance on the IHDP dataset with $N^{\rm{s}}=6$.
Compared with the performance of the baseline methods in Table~\ref{tab:mse},
our proposed method with all the learners achieved good performance,
which indicates the effectiveness of closed-form solvers
for task adaptation.


Table~\ref{tab:variant_gp} shows the average PEHE
when using Gaussian processes (GPs) for task adaptation
in the proposed method as described in Section~\ref{sec:gp}.
Linear is the proposed method using linear models for task adaptation
as described in Section~\ref{sec:solver},
which corresponds to Ours in Table~\ref{tab:mse}.
The performance with GPs was comparable to that with linear models.
This result indicates that the proposed method also works well when GPs
are used for task adaptation instead of linear models.

To see how computationally efficient these closed-form solvers are, we evaluated
the computation time that is required to perform meta-learning
using computers with CentOS, Xeon Gold 6130 2.10GHz CPU, and 256GB memory.
Table~\ref{tab:time} shows the results.
Compared with the methods that perform iterative optimization for adaptation (i.e., DR-ML and Meta-CI),
those using closed-form solvers (i.e., Ours and DR-CFS)
needed much shorter times, implying that adapting task-specific parameters with closed-form solvers
is far more computationally efficient.

\begin{table}[t!]
  \centering
  \caption{Average PEHE and its standard error. $N_{0}^{\rm{s}}$ and $N_{1}^{\rm{s}}$ are number of untreated and treated support instances. Values in bold are not statistically different at 5\% level from best performing method in each dataset by a paired t-test.}
  \label{tab:mse}
  
  \begin{tabular}{lrrr}
  \multicolumn{4}{c}{(a) Synth}\\
        \hline
$N^{\rm{s}}$ & 6 & 10 & 14\\        
        \hline
Ours & {\bf 4.941 $\pm$ 0.133} & {\bf 4.826 $\pm$ 0.121} & {\bf 4.736 $\pm$ 0.128} \\
DR-CFS & 5.767 $\pm$ 0.165 & 6.246 $\pm$ 0.155 & 6.547 $\pm$ 0.179 \\
DR-ML & 5.311 $\pm$ 0.160 & 5.322 $\pm$ 0.157 & 5.318 $\pm$ 0.158 \\
Meta-CI & 5.295 $\pm$ 0.161 & 5.302 $\pm$ 0.158 & 5.304 $\pm$ 0.158 \\
MT & 11.463 $\pm$ 0.391 & 13.922 $\pm$ 0.400 & 16.143 $\pm$ 0.508 \\
TL & 11.738 $\pm$ 0.383 & 11.643 $\pm$ 0.332 & 12.710 $\pm$ 0.284 \\
SL & {\bf 5.001 $\pm$ 0.142} & {\bf 4.879 $\pm$ 0.123} & 4.878 $\pm$ 0.130 \\
XL & 11.001 $\pm$ 0.374 & 10.065 $\pm$ 0.355 & 10.355 $\pm$ 0.286 \\
DRL & 175.808 $\pm$ 30.994 & 348.802 $\pm$ 59.052 & 2942.049 $\pm$ 1494.359 \\
CF & 13.470 $\pm$ 0.572 & 9.735 $\pm$ 0.516 & 7.941 $\pm$ 0.310 \\
Mean & 5.580 $\pm$ 0.193 & 5.535 $\pm$ 0.184 & 5.469 $\pm$ 0.171\\
\hline
    \end{tabular}
    
  \begin{tabular}{lrrr}
  \multicolumn{4}{c}{(b) IHDP}\\
        \hline
$N^{\rm{s}}$ & 6 & 10 & 14\\        
        \hline
Ours & {\bf 1.356 $\pm$ 0.026} & {\bf 1.255 $\pm$ 0.028} & {\bf 1.205 $\pm$ 0.025} \\
DR-CFS & 2.907 $\pm$ 0.095 & 3.204 $\pm$ 0.088 & 3.460 $\pm$ 0.089 \\
DR-ML & 2.731 $\pm$ 0.050 & 2.720 $\pm$ 0.050 & 2.714 $\pm$ 0.048 \\
Meta-CI & 2.783 $\pm$ 0.044 & 2.765 $\pm$ 0.044 & 2.761 $\pm$ 0.044 \\
MT & 4.134 $\pm$ 0.109 & 4.788 $\pm$ 0.132 & 5.500 $\pm$ 0.413 \\
TL & 3.026 $\pm$ 0.105 & 3.373 $\pm$ 0.109 & 3.695 $\pm$ 0.101 \\
SL & 3.460 $\pm$ 0.103 & 2.303 $\pm$ 0.072 & 1.981 $\pm$ 0.063 \\
XL & 2.563 $\pm$ 0.068 & 2.738 $\pm$ 0.081 & 3.035 $\pm$ 0.088 \\
DRL & 207.866 $\pm$ 34.684 & 409.598 $\pm$ 100.264 & 745.870 $\pm$ 173.200 \\
CF & 3.193 $\pm$ 0.146 & 2.028 $\pm$ 0.094 & 1.743 $\pm$ 0.066 \\
Mean & 2.747 $\pm$ 0.050 & 2.737 $\pm$ 0.048 & 2.716 $\pm$ 0.046\\
\hline
      \end{tabular}
\end{table}

\begin{table}[t!]
  \centering
  \caption{Average PEHE and its standard error of proposed method with DR-, RA-, and Plugin-Learners. Values in bold are not statistically different at 5\% level from best performing method in each dataset by a paired t-test.}
  \label{tab:variant}
  \begin{tabular}{lrrr}
  \multicolumn{4}{c}{(a) Synth}\\
        \hline
$N^{\rm{s}}$ & 6 & 10 & 14\\
\hline
w/ DR & {\bf 4.941 $\pm$ 0.133} & {\bf 4.826 $\pm$ 0.121} & {\bf 4.736 $\pm$ 0.128}\\
w/ RA & {\bf 4.950 $\pm$ 0.141} & {\bf 4.804 $\pm$ 0.124} & {\bf 4.673 $\pm$ 0.129}\\ 
w/ Plugin & {\bf 4.939 $\pm$ 0.135} & {\bf 4.787 $\pm$ 0.122} & 4.714 $\pm$ 0.135\\
\hline
\end{tabular}

      \begin{tabular}{lrrr}
  \multicolumn{4}{c}{(b) IHDP}\\    
        \hline
$N^{\rm{s}}$ & 6 & 10 & 14\\
\hline
w/ DR & {\bf 1.356 $\pm$ 0.026} & {\bf 1.255 $\pm$ 0.028} & {\bf 1.205 $\pm$ 0.025}\\
w/ RA & 1.462 $\pm$ 0.033 & {\bf 1.279 $\pm$ 0.028} & {\bf 1.185 $\pm$ 0.030}\\
w/ Plugin & 1.532 $\pm$ 0.039 & 1.312 $\pm$ 0.036 & {\bf 1.205 $\pm$ 0.029}\\
\hline
\end{tabular}
\end{table}

\begin{table}[t!]
  \centering
  \caption{Average PEHE and its standard error of proposed method with linear models (Linear) and Gaussian processes (GP) for task adaptation. Values in bold are not statistically different at 5\% level from best performing method in each dataset by a paired t-test.}
  \label{tab:variant_gp}
  \begin{tabular}{lrrr}
  \multicolumn{4}{c}{(a) Synth}\\
        \hline
$N^{\rm{s}}$ & 6 & 10 & 14\\
\hline
Linear & {\bf 4.941 $\pm$ 0.133} & {\bf 4.826 $\pm$ 0.121} & {\bf 4.736 $\pm$ 0.128}\\
GP & {\bf 5.102 $\pm$ 0.160} & {\bf 4.888 $\pm$ 0.163} & {\bf 4.730 $\pm$ 0.132}\\
\hline
\end{tabular}

\begin{tabular}{lrrr}
  \multicolumn{4}{c}{(b) IHDP}\\    
        \hline
$N^{\rm{s}}$ & 6 & 10 & 14\\
\hline
Linear & {\bf 1.356 $\pm$ 0.026} & {\bf 1.255 $\pm$ 0.028} & {\bf 1.205 $\pm$ 0.025}\\
GP & {\bf 1.335 $\pm$ 0.029} & {\bf 1.284 $\pm$ 0.042} & {\bf 1.213 $\pm$ 0.030}\\
\hline
\end{tabular}
\end{table}

\begin{table}[t!]
  \centering
  \caption{Computation time in seconds for meta-training with $N^{\rm{s}}=6$.}
  \label{tab:time}
  \begin{tabular}{lrrrr}
    \hline
    & Ours & DR-CFS & DR-ML & Meta-CI \\
    \hline
    Synth & 1402.7 & 1454.6 & 2009.5 & 2634.1 \\
    IHDP & 442.3 & 489.9 & 1003.8 & 1751.9 \\
    \hline
  \end{tabular}
\end{table}

\section{Conclusion}

We proposed a meta-learning framework
for improving the CATE estimation performance on a few observational
data. We formulated our CATE estimation models using
neural networks with task-shared and task-specific parameters
and developed an effective strategy with differentiable closed-form solvers for adapting task-specific parameters.
The task-shared parameters are trained such that the CATE estimation with a few data approximates that with large data.
We experimentally show that our framework achieved better CATE estimation performance than the existing methods.
A future work direction is to develop a meta-learning
approach in the presence of unobserved confounders.

\bibliographystyle{abbrv}
\bibliography{ml2023cate}

\end{document}